%% file: main.tex
\documentclass{article}

\usepackage{microtype}
\usepackage{graphicx}
\usepackage{booktabs}
\usepackage{xcolor}

\usepackage{nicefrac}      
\usepackage{microtype}     
\usepackage{graphicx}
\usepackage{subcaption}
\usepackage{amsfonts}      
\usepackage{amssymb}
\usepackage{amsmath}
\usepackage{amsthm}
\usepackage{wrapfig}
\usepackage{enumitem}
\usepackage{bm}
\usepackage{pgfplots}
\usepackage{custom}

\usepackage{hyperref}
\usepackage{cleveref}

\usepackage[accepted]{icml2021}

\icmltitlerunning{Solving Inverse Problems with a Flow-based Noise Model}

\begin{document}

\twocolumn[
\icmltitle{Solving Inverse Problems with a Flow-based Noise Model}
\icmlsetsymbol{equal}{*}

\begin{icmlauthorlist}
\icmlauthor{Jay Whang}{utcs}
\icmlauthor{Qi Lei}{princeton}
\icmlauthor{Alexandros G. Dimakis}{utece}
\end{icmlauthorlist}

\icmlaffiliation{utcs}{Dept. of Computer Science, UT Austin, TX, USA}
\icmlaffiliation{utece}{Dept. of Electrical and Computer Engineering, UT Austin, TX, USA}
\icmlaffiliation{princeton}{Dept. of Electrical and Computer Engineering, Princeton University, NJ, USA}

\icmlcorrespondingauthor{Jay Whang}{\mbox{\texttt{jaywhang@utexas.edu}}}

\icmlkeywords{inverse problem, normalizing flow}

\vskip 0.3in
]

\printAffiliationsAndNotice{}

\begin{abstract}
We study image inverse problems with a normalizing flow prior. Our formulation views the solution as the maximum a posteriori estimate of the image conditioned on the measurements. This formulation allows us to use noise models with arbitrary dependencies as well as non-linear forward operators. We empirically validate the efficacy of our method on various inverse problems, including compressed sensing with quantized measurements and denoising with highly structured noise patterns. We also present initial theoretical recovery guarantees for solving inverse problems with a flow prior.
\end{abstract}

\section{Introduction}

\label{sec:intro}
Inverse problems seek to reconstruct an unknown signal from observations (or \textit{measurements}), which are produced by some process that transforms the original signal.
Because such processes are often lossy and noisy,
inverse problems are typically formulated as reconstructing $\bx$ from its measurements 
\begin{equation}
    \xtil = f(\bx) + \bdelta
    \label{eqn:inverse_problem}
\end{equation}
where $f$ is a known deterministic \textit{forward operator} and $\bdelta$ is an additive noise which may have a complex structure itself.
An impressively wide range of applications can be posed under this formulation with an appropriate choice of $f$ and $\bdelta$, such as compressed sensing~\citep{candes2006stable,donoho2006compressed}, computed tomography \cite{Chen2008PriorIC}, magnetic resonance imaging (MRI)~\citep{lustig2007sparse}, and phase retrieval~\citep{candes2015phase,candes2015Wirtinger}.

In general, for a non-invertible forward operator $f$, there can be potentially infinitely many signals that match given observations.
Thus the recovery algorithm must critically rely on \textit{a priori} knowledge about the original signal to find the most plausible solution among them.
Sparsity has classically been a very influential structural prior for various inverse problems~\citep{candes2006stable,donoho2006compressed,baraniuk2010model}.
Alternatively, recent approaches introduced deep generative models as a powerful signal prior, showing significant gains in reconstruction quality compared to sparsity priors~\citep{bora2017compressed,PaulHandFlow,van2018compressed,menon2020pulse}.

However, most existing methods assume the setting of Gaussian noise model and are unsuitable for structured and correlated noise, making them less applicable in many real-world scenarios. For instance, when an image is ruined by hand-drawn scribbles or a piece of audio track is overlayed with human whispering, the noise to be removed follows a very complex distribution.  These settings deviate significantly from Gaussian noise settings, yet they are much more realistic and deserve more attention. In this paper, we propose to use a normalizing flow model to represent the noise distribution and derive principled methods for solving inference problems under this general noise model.

\input{figures/fig_intro.tex}

\textbf{Contributions.}

\begin{itemize}
    \item We present a general formulation for obtaining maximum a posteriori (MAP) reconstructions for dependent noise and general forward operators. Notably, our method can leverage deep generative models for both the original image and the noise.
    \item We extend our framework to the general setting where the signal prior is given as a latent-variable model, for which likelihood evaluation is intractable.  The resulting formulation presents a unified view of existing approaches based on GAN, VAE, and flow priors.
    \item We empirically show that our method achieves excellent reconstruction in the presence of noise with various complex and dependent structures. Specifically, we demonstrate the efficacy of our method on various inverse problems with structured noise and non-linear forward operators.
    \item We provide the initial theoretical characterization of likelihood-based priors for image denoising. Specifically, we show a reconstruction error bound that depends on the local concavity of the log-likelihood function.  
\end{itemize}

\section{Background}

\subsection{Normalizing Flow Models}
Normalizing flow models are a class of likelihood-based generative models that represent complex distributions by transforming a simple distribution (such as standard Gaussian) through an invertible mapping  \citep{tabak2013family}.  Compared to other types of generative models, flow models are computationally flexible in that they provide efficient sampling, inversion, and likelihood estimation~\citep[and references therein]{papamakarios2019normalizing}.

Concretely, given a differentiable invertible mapping $G: \R^n \to \R^n$, the samples $\bx$ from this model are generated via
$\bz \sim \ptrue(\bz), \bx = G(\bz)$.
Since $G$ is invertible, change of variables formula allows us to compute the log-density of $\bx$:
\begin{equation}
\label{eqn:flow_logp}
\log p(\bx) = \log p(\bz) + \log \abs{\det J_{G^{-1}}(\bx)},
\end{equation}
where $J_{G^{-1}}(\bx)$ is the Jacobian of $G^{-1}$ evaluated at $\bx$.
Since $\log p(\bz)$ is a simple distribution, computing the likelihood at any point $\bx$ is straightforward as long as $G^{-1}$ and the log-determinant term can be efficiently evaluated.

Notably, when a flow model is used as the prior for an inverse problem, the invertibility of $G$ guarantees that it has an unrestricted range. Thus the recovered signal can represent images that are out-of-distribution, albeit at lower probability.
This is a key distinction from a GAN-based prior, whose generator has a restricted range and can only generate samples from the distribution it was trained on.
As pointed out by \citet{PaulHandFlow} and also shown below in our experiments, this leads to performance benefits on out-of-distribution examples.

\subsection{Inverse Problems with a Generative Prior}
We briefly review the existing literature on the application of deep generative models to inverse problems.
While vast literature exists on compressed sensing and other inverse problems,
the idea of replacing the classical sparsity-based prior \citep{candes2006stable,donoho2006compressed} with a neural network was introduced relatively recently. 
In their pioneering work, \citet{bora2017compressed} proposed to use the generator from a pre-trained GAN or a VAE \citep{goodfellow2014generative,kingma2013auto} as the prior for compressed sensing.  This led to a substantial gain in reconstruction quality compared to classical methods, particularly at a small number of measurements. 

Following this work, numerous studies have investigated different ways to utilize various neural network architectures for inverse problems \citep{mardani2018neural,heckel2018deep,mixon2018sunlayer, pandit2019asymptotics,lucas2018using,shah2018solving,liu2020information,kabkab2018task,lei2019inverting,mousavi2018data, raj2019gan,sun2019block}.  One straightforward extension of \citep{bora2017compressed} proposes to expand the range of the pre-trained generator by allowing sparse deviations \citep{dhar2018modeling}.
Similarly, \citet{shah2018solving} proposed another algorithm based on projected gradient descent with convergence guarantees.
\citet{van2018compressed} showed that an \textit{untrained} convolutional neural network can be used as a prior for imaging tasks based on Deep Image Prior by \citet{ulyanov2018deep}.

More recently, \citet{wu2019deepcs} applied techniques from meta-learning to improve the reconstruction speed, 
and \citet{ardizzone2018} showed that by modeling the forward process with a flow model, one can \textit{implicitly} learn the inverse process through the invertibility of the model.
\citet{PaulHandFlow} proposed to replace the GAN prior of \citep{bora2017compressed} with a normalizing flow model and reported excellent reconstruction performance, especially on out-of-distribution images.

\section{Our Method}

\subsection{Notations and Setup}
We use bold lower-case variables to denote vectors, $\|\cdot\|$ to denote $\ell_2$ norm, and $\odot$ to denote element-wise multiplication.
We also assume that we are given a pre-trained latent-variable generative model $\ptrue(\bx)$ that we can efficiently sample from.  Importantly, we assume the access to a noise distribution $\pnoise$ parametrized as a normalizing flow, which itself can be an arbitrarily complex, pre-trained distribution. We let $f$ denote the deterministic and differentiable forward operator for our measurement process.  
Thus an observation is generated via $\xtil = f(\bx) + \bdelta$ where $\bx \sim \ptrue(\bx)$ and $\bdelta \sim \pnoise(\bdelta)$.

Note that while $f$ and $\pnoise$ are known, they cannot be treated as fixed across different examples,~e.g., in compressed sensing, the measurement matrix is random and thus only known at the time of observation. This precludes the use of end-to-end training methods that require having a fixed forward operator.

\subsection{MAP Formulation}

When the likelihood under the prior $\ptrue(\bx)$ can be computed efficiently (e.g. when it is a flow model), we can pose the inverse problem as a MAP estimation task. Concretely, for a given observation $\xtil$, we wish to recover $\bx$ as the MAP estimate of the conditional distribution $\ptrue(\bx \vert \xtil)$:
\begin{align*}
    &\argmax_\bx \log p(\bx \vert \xtil) \\
    &= \argmax_\bx \brac{\log p(\xtil \vert \bx) + \log \ptrue(\bx) - \log p(\xtil)} \\
    &\stackeq{(1)} \argmax_\bx \brac{\log \pnoise(\xtil-f(\bx)) + \log \ptrue(\bx)} \\
    &\triangleq \argmin_\bx \Lmap(\bx;\xtil),
\end{align*}
where
\begin{equation}
  \label{eqn:map_loss}
  \Lmap(\bx;\xtil) = -\log\pnoise(\xtil-f(\bx)) - \log \ptrue(\bx).
\end{equation}
Note that in (1) we drop the marginal density $\log p(\xtil)$ as it is constant and rewrite $p(\xtil \vert \bx)$ as $\pnoise(\xtil-f(\bx))$.

Recalling that the generative procedure for the flow model is $\bz \sim \N(\bzero,I), \bx=G(\bz)$,
we arrive at the following loss:
\begin{equation}
\label{eqn:loss_g}
\Lmap(\bz; \xtil) \triangleq -\log\pnoise(\xtil-f(G(\bz))) - \log \pmodel(G(\bz))
\end{equation}
The invertibility of $G$ allows us to minimize the above loss with respect to either $\bz$ or $\bx$:
\begin{align*}
&\argmin_\bz \,\,\LG(\bz; \xtil) \\
&= \argmin_\bz \brac{-\log\pnoise\paren{\xtil-f(G(\bz))} - \log \pmodel(G(\bz))} \\
&= \argmin_\bx \brac{-\log\pnoise\paren{\xtil-f(\bx)} - \log \pmodel(\bx))} \\
&= \argmin_\bx \,\,\LG(\bx; \xtil)
\end{align*}
We have experimented with optimizing the loss both in image space $\bx$ and latent space $\bz$, and found that the latter achieved better performance across almost all experiments. 
Since the above optimization objective is differentiable, any gradient-based optimizer can be used to find the minimizer approximately. In practice, even with an imperfect model and approximate optimization,
we observe that our approach performs well across a wide range of tasks, as shown in the experimental results below.

\subsection{MLE Formulation}

When the signal prior does not provide tractable likelihood evaluation (e.g. for the case of GAN and VAE), we view the problem as a maximum-likelihood estimation under the \textit{noise model}.  Thus we attempt to find the signal that maximizes noise likelihood within the support of $\ptrue(\bx)$ and arrive at a similar, but different loss:

\begin{align*}
    &\argmax_{\bx \in \textrm{supp } p(\bx)} \log \pnoise(\by - f(\bx)) \\
    &= \argmax_{\bz} \log \pnoise(\by - f(G(\bz))) \\
    &\triangleq \argmin_\bz \Lmle(\bz;\xtil),
\end{align*}
where
\begin{equation}
\label{eqn:loss_g_mle}
\Lmle(\bz; \xtil) \triangleq -\log\pnoise(\xtil-f(G(\bz))).
\end{equation}

\subsection{Prior Work}

In \citep{bora2017compressed}, the authors proposed to use a deep generative prior for the inverse problem, but the choice of models was restricted to GANs and VAEs with explicit low-dimensional prior. Subsequently \citet{PaulHandFlow} generalized this paradigm using Flow-based models. We describe here the methods proposed in those papers in detail. Importantly, we show that their approaches are special cases of our MAP/MLE formulations under Gaussian noise assumptions. Furthermore, note that both papers considered linear inverse problems, so they correspond to the case where $f(\bx) = A\bx$ under our notation.

\textbf{GAN Prior:}
\citep{bora2017compressed} considers the following loss:
\begin{equation}
    \Lbora(\bz;\xtil) = \normsq{\xtil - AG(\bz)} + \lambda \normsq{z},
\end{equation}
which tries to project the input $\xtil$ onto the range of the generator $G$ with $\ell_2$ regularization on the latent variable.
Aside from the regularization term, this corresponds exactly to our MLE loss for a Gaussian $\pnoise$.  While \citep{bora2017compressed} motivated this objective as a projection on the range of $G$, our approach reveals a probabilistic interpretation based on the MLE objective for the noise.

\textbf{Flow Prior:}
\citep{PaulHandFlow} replaces the GAN prior of \citep{bora2017compressed} with a  flow model.
In that paper, the authors consider the objective below that tries to simultaneously match the observation and maximize the likelihood of the reconstruction under the model:
\begin{equation}
\label{eqn:L_hand_general}
    L(\bz; \xtil) = \normsq{\xtil - AG(\bz)} - \gamma \log \pmodel(\bx),
\end{equation}
for some hyperparameter $\gamma > 0$.
This loss is a special case of our MAP loss for isotropic Gaussian noise $\bdelta \sim \N(\bzero,\gamma\,I)$,
since the log-density of $\bdelta$ becomes $\log \pnoise(\bdelta) = -\frac{1}{2\gamma}\normsq{\bdelta} - C$ for a constant $C$.
However, \citet{PaulHandFlow} report that due to optimization difficulty, they found the following proxy loss to perform better in experiments:
\begin{equation}
    \Lhand(\bz;\xtil) = \normsq{\xtil-AG(\bz)} + \gamma \norm{\bz}.
\end{equation}
This is again related to a specific instance of our loss when the flow model is volume-preserving (i.e., the log-determinant term is constant). Continuing from \cref{eqn:flow_logp}:
This allows us to recover the $\ell_2$-regularized version of $\Lhand$:

We reiterate that both of the aforementioned objectives are special cases of our formulation for the case of zero-mean isotropic Gaussian noise. Thus, we expect our method to handle non-Gaussian noise better and experimentally confirm that our approach leads to better reconstruction performance for noises with nonzero mean or conditional dependence across different pixel locations.

\textbf{Connections to Blind Source Separation:}
Here we focus on the connection between our formulation and blind source separation.  For the denoising case where the forward operator is identity, we see that our observation is simply the sum of two random variables $\by = \bx + \bdelta$.  Given two flow-based priors (one for each of $\bx$ and $\bdelta$), the task of extracting $\bx$ from $\by$ thus becomes a blind source separation problem with two sources.
While rich literature exists for various source separation problems \citep{hu2017deeplearning,subakan2018generative,wang2018supervised,hoshen2019towards}, two recent studies are particularly relevant to our setting as they make use of a neural network prior.

In Double-DIP, \citet{gandelsman2019double} utilize Deep Image Prior \citep{ulyanov2018deep} as a signal prior to performing blind source separation from multiple mixtures.  This work differs from ours in that we focus on a single-mixture setting with pre-trained signal priors.  Our use of pre-trained priors is a key distinction since DIP is untrained and may not apply to other modalities and datasets. In contrast, our method is applicable as long as we are able to train a deep generative prior for the signal and the noise.

In \citep{jayaram2020source}, the authors use a flow-based prior \citep{kingma2018glow} for blind source separation.  Unlike our approach, however, they sample from the posterior using Langevin dynamics \citep{welling2011bayesian,neal2011mcmc}.  The authors use simulated annealing to speed up mixing, and this approach would in theory be able to sample from the correct posterior asymptotically.  The advantage of our approach is that it is generally faster (as it avoids costly MCMC procedure), and it can be applied to non-likelihood-based priors for the signal $\bx$.

\section{Theoretical Analysis}
This section provides some theoretical analysis of our approach in denoising problems with a flow-based prior. Unlike most prior work, we take a probabilistic approach and avoid making specific structural assumptions on the signal we want to recover, such as sparsity or being generated from a low-dimensional Gaussian prior. 

For denoising, we show that better likelihood estimates lead to lower reconstruction error.
Note that while our experiments employed flow models, our results apply to any likelihood-based generative model. The detailed proof is included in the appendix.

\subsection{Recovery Guarantee for Denoising}

Suppose we observe $ \xtil=\bx^*+\bdelta$ with Gaussian noise $\bdelta \sim \N(\bzero, \sigma^2 I)$ with $\norm{\bdelta} = r$.
We perform MAP inference by minimizing the following loss with gradient descent:
\begin{align}
\notag 
\Lmap(\bx) = & -\log\pnoise(\xtil-\bx) -\log p(\bx)\\
\label{eqn:denoising}
= & \frac{1}{2\sigma^2}\normsq{\xtil-\bx} + q(\bx),
\end{align}
where we write $q(\bx) \triangleq -\log p(\bx)$ for notational convenience.
Notice that the image we wish to recover is a natural image with high probability rather than an arbitrary one, and reconstruction is not expected to succeed for the latter case. Thus we consider the case where the ground truth image $\bx^*$ is a local maximum of $p$. 
\begin{theorem}
\label{thm:denoising}
Let $\bx^*$ be a local optimum of the model $p(\bx)$ and $\xtil=\bx^*+\bdelta$ be the noisy observation.   
Assume that $q$ satisfies local strong convexity within the ball around $\bx^*$ defined as $B_r^d(\bx^*) \triangleq \set{\bx \in \R^d : \|\bx-\bx^*\|\leq r}$, i.e. the Hessian of $q$ satisfies $H_q(\bx)\succeq \mu I$ $\forall \bx \in B_r^d(\bx^*)$ for for some $\mu>0$.
Then gradient descent starting from $\xtil$ on the loss function \eqref{eqn:denoising} converges to $\bar{\bx}$, a local minimizer of $\Lmap(\bx)$, that satisfies:
\begin{equation*}
    \|\bar{\bx}-\bx^*\|\leq \frac{1}{\mu\sigma^2+1}\|\bdelta\|
\end{equation*}
\end{theorem}
Even though the theorem is relatively straightforward, it still serves as some initial understanding of the denoising task under a likelihood-based prior. It sheds light on how the reconstruction is affected by the structure of the probabilistic model, and the likelihood of the natural signal one wants to recover. This theorem shows that a well-conditioned model with large $\mu$ leads to better denoising and confirms that our MAP formulation encourages reconstructions with high density. Thus, the maximum-likelihood training objective is directly aligned with better denoising performance.

\section{Experiments}

\input{figures/fig_mnist_many_samples_both.tex}

Our experiments are designed to test how well our MAP/MLE formulation performs in practice as we deviate from the commonly studied setting of linear inverse problems with Gaussian noise.  Specifically, we focus on two aspects: 
(1) complex noise with dependencies and (2) non-linear forward operator.
For all our experiments, we quantitatively evaluate each method by reporting peak signal-to-noise ratio (PSNR). We also visually inspect sample reconstructions for qualitative assessment.

\textbf{Models:}
We trained multi-scale RealNVP models on two image datasets MNIST and CelebA-HQ \citep{lecun1998gradient,liu2015faceattributes}.
Due to computational constraints, all experiments were done on 100 randomly-selected
images (1000 for MNIST) from the test set, as well as out-of-distribution images.  We additionally train a DCGAN on the CelebA-HQ dataset for MLE experiments as well as \citep{bora2017compressed}.
A detailed description of the datasets, models, and hyperparameters are provided in the appendix.

\textbf{Baseline Methods:}
We compare our approach to the methods of \citep{bora2017compressed} and \citep{PaulHandFlow}, as they are two recently proposed approaches that use deep generative prior on inverse problems.
Depending on the task, we also compare against BM3D \citep{dabov2006image}, a popular off-the-shelf image denoising algorithm, and LASSO \citep{tibshirani1996regression} with Discrete Cosine Transform (DCT) basis as appropriate.  Note that for the 1-bit compressed sensing experiment, most existing techniques do not apply, since our task involves quantization as well as non-Gaussian noise. 

We point out that the baselines methods are not designed to make use of the noise distribution, whereas our method does utilize it.
Thus, the experiments are not meant to be taken as direct comparisons, but rather as empirical evidence that the MAP formulation indeed benefits from the knowledge of noise structure.

\textbf{Smoothing Parameter:}
Since our objective \Cref{eqn:loss_g} depends on the density $\pmodel(\bx)$ given by the flow model, our recovery of $\bx$ depends heavily on the quality of density estimates from the model. 
Unfortunately, likelihood-based models exhibit counter-intuitive properties, such as assigning higher density on out-of-distribution examples or random noise over in-distribution examples. \citep{nalisnick2018deep,choi2018waic,hendrycks2018benchmarking,nalisnick2019detecting}.
We empirically observe such behavior from our models as well.
To remedy this, we use a smoothed version of the model density $\pmodel(x)^\beta$ where $\beta \ge 0$ is the \textit{smoothing parameter}.  Since the two extremes $\beta=0$ and $\beta=\infty$ correspond to only using the noise density and the model density, respectively, $\beta$ controls the degree to which we rely on $\pmodel$.
Thus the loss we minimize becomes
\begin{equation}
    \label{eqn:loss_g_beta}
    \LG(\bz;\xtil,\beta) = -\log \pnoise(\xtil - f(G(\bz))) - \beta \log \pmodel(G(\bz))
\end{equation}

\subsection{Results}
We tested our methods on denoising and compressed sensing tasks that involve various structured, non-Gaussian noises as well as a non-linear forward operator. Note that many existing methods cannot be applied in this setting as they are designed for linear inverse problems and a specific noise model. While specialized algorithms do exist (e.g., for quantized compressed sensing), we note that our method is general and can be applied in a wide range of settings without modification.

\subsubsection{Denoising MNIST Digits}

The measurement process is $\xtil = 0.5 \cdot \bx + \bdelta_{\MNIST}$,
where $\bdelta_{\MNIST}$ represents MNIST digits added at different locations and color channels. Each digit itself comes from a flow model trained on the MNIST dataset.
As shown in \Cref{fig:mnist_many_samples}, our method successfully removes MNIST noise in both MAP and MLE settings.  Recall that we use the same GAN-based prior for ``Ours (MLE)'' and \citep{bora2017compressed}.
The difference in the reconstruction quality between these two methods in \Cref{fig:mnist_psnr} confirms that even for non-likelihood-based priors (e.g., GAN and VAE), an accurate noise model is critical to accurate signal recovery.
While the method of \citet{bora2017compressed} manages to remove MNIST digits, we note that this is because its outputs are forced to be in the range of the DCGAN.

\input{figures/fig_mnist_psnr}

For the rest of the experiments, we focus on the MAP formulation as it generally outperforms the MLE formulation. We posit that this is due to the restricted range of our DCGAN used for experiments.

\subsubsection{Noisy Compressed Sensing}

Now we consider the measurement process $\xtil = A\bx + \bdelta_{\text{sine}}$ where $A \in \R^{m \times d}$ is a random Gaussian measurement matrix and $\bdelta$ has positive mean with variance that follows a sinusoidal pattern. Specifically, the $k$-th pixel of the noise has standard deviation $\sigma_k \propto \exp\paren{\sin(\frac{2\pi k}{16})}$ normalized to have the maximum variance of 1.

\Cref{fig:cs_noise_psnr} shows that our method is able to make better use of additional measurements. Interestingly in \Cref{fig:cs_noise_samples}, all three methods with deep generative prior produced plausible human faces.  However, the reconstructions from \citet{PaulHandFlow} and \citet{bora2017compressed} significantly differ from the ground truth images.  We posit that this is due to the implicit Gaussian noise assumption made by the two methods, again showing the benefits of explicitly incorporating the knowledge of noise distribution.

\input{figures/fig_cs_noise_psnr_and_samples.tex}

\subsubsection{Removing \sinusoidal Noise}

We consider another denoising task with observation $\xtil = \bx + \bdelta_{\text{sine}}$.
This is the 2-dimensional version of periodic noise, where the noise variance for all pixels within the $k$-th row follows $\sigma_k$ from above.

From \Cref{fig:sinusoidal_samples_both} and \Cref{fig:sinusoidal_psnr}, we see that baseline methods do not perform well even though the noise is simply Gaussian at each pixel.
This reemphasizes an important point: without an understanding of the \textit{structure} of the noise, algorithms designed to handle Gaussian noise have difficulty removing them when we introduce variability across different pixel locations.
\input{figures/fig_sinusoidal_samples_both.tex}
\input{figures/fig_sinusoidal_and_hparam_sensitivity}

\subsubsection{Noisy 1-bit Compressed Sensing}

This task considers a combination of a non-linear forward operator as well as a non-Gaussian noise.
The measurement process is $\xtil = \text{sign}(A\bx) + \bdelta_{\text{sine}}$, identical to noisy compressed sensing except with the sign function.
This is the most extreme case of quantized compressed sensing, since $\xtil$ only contains the (noisy) sign $\set{+1,-1}$ of the measurements.
Because the gradient of sign function is zero everywhere, we use Straight-Through Estimator \cite{bengio2013estimating} for backpropagation.
See \Cref{fig:cs_1bit_measurements} and \Cref{fig:cs_1bit_samples} for a comparison of our method to the baselines at varying numbers of measurements.

\input{figures/fig_cs_1bit_psnr_and_samples.tex}

\subsubsection{Sensitivity to Hyperparameters}
We also observe that our method is more robust to hyperparameter choices than \citep{PaulHandFlow}, as shown in \Cref{fig:cs_noise_hparam_sensitivity}.
This may come as a surprise, given that our objective has an additional log-determinant term.
We speculate that this is due to the regularization term in $\Lhand(\bz;\xtil) = \normsq{\xtil-f(G(\bz))} + \gamma \norm{\bz}$.

It is known that samples from $d$-dimensional isotropic Gaussian concentrate around a thin ``shell'' around the sphere of radius $\sqrt{d}$.  This suggests that the range of $\norm{\bz}$ corresponding to natural images may be small.  Thus, forcing the latent variable $\bz$ to have a small norm without taking the log-determinant term into account could lead to a sudden drop in the reconstruction quality.

\section{Conclusion}

We propose a novel method to solve inverse problems for general differentiable forward operators and structured noise. 
Our method generalizes that of~\citep{PaulHandFlow} to arbitrary differentiable forward operators and non-Gaussian noise distributions. The power of our approach stems from the flexibility of invertible generative models, which can be combined in a modular way to solve MAP inverse problems in very general settings, as we demonstrate. 

For future work, it would be interesting to consider extending our method to blind source separation problems~\citep{amari1996new}, as the noise and signal considered in our setting are actually exchangeable for the MAP case. Furthermore, one may investigate the applicability of the MLE formulation with an even more general family of generative models such as energy-based models \citep{Gao_2020_CVPR} and score networks \citep{song2019generative}.
On the theoretical side, one central question that remains open is to analyze the optimization problem we formulated. In this paper, we empirically minimize this loss using gradient descent, but some theoretical guarantees would be desirable, possibly under assumptions, e.g. random weights following the framework of~\cite{hand2017global}.

\section*{Acknowledgements}
This research has been supported by NSF Grants CCF 1934932, AF 1901292,
2008710, 2019844 the NSF IFML 2019844 award as well as research gifts by Western Digital, WNCG, and MLL, computing resources from TACC and the Archie Straiton Fellowship, and NSF 2030859 for the Computing Research Association for the CIFellows Project.

\clearpage
\bibliography{main}
\bibliographystyle{icml2021}

\clearpage
\appendix
\input{appendix}

\end{document}

%% file: figures/fig_intro.tex
\begin{figure}[!b]
\centering
\includegraphics[width=0.95\linewidth]{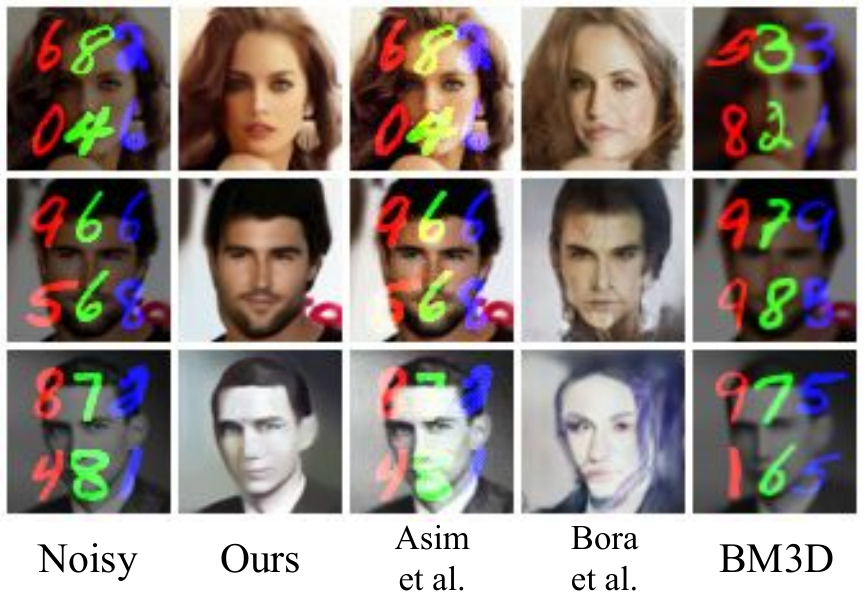}
\vspace{-0.5em}
\caption{
Result of denoising MNIST digits. The first column contains noisy observations, and subsequent columns contain reconstructions.
}
\label{fig:intro}
\end{figure}

%% file: figures/fig_mnist_many_samples_both.tex
\begin{figure*}[!ht]
\centering
\includegraphics[width=0.9\linewidth]{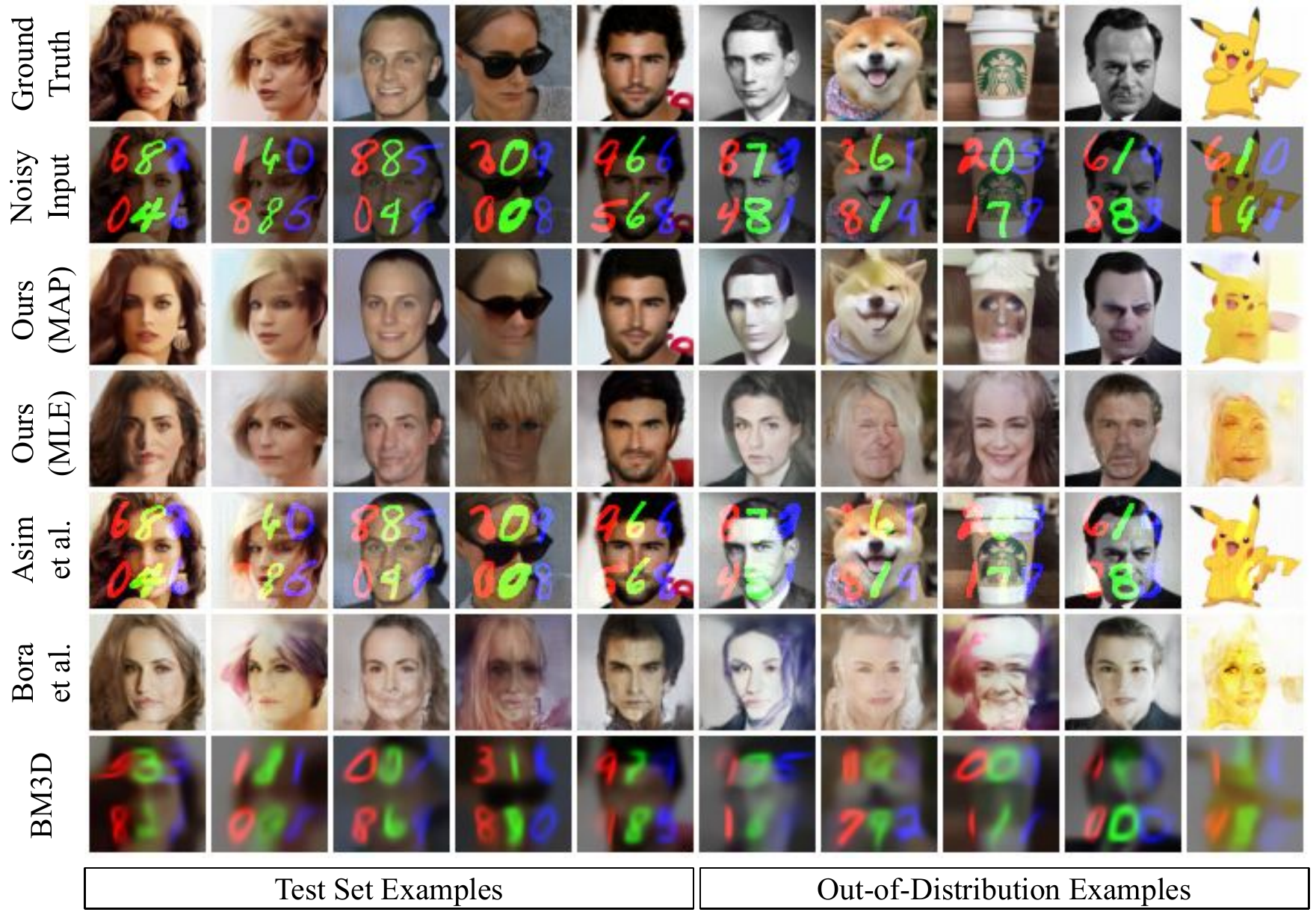}
\caption{Result of removing MNIST noise from CelebA-HQ faces.  Notice that without any understanding of the complex noise structure, baseline methods fail to produce good reconstructions.}
\label{fig:mnist_many_samples}
\end{figure*}

%% file: figures/fig_mnist_psnr.tex
\begin{figure}[!h]
\centering
\includegraphics[width=\linewidth]{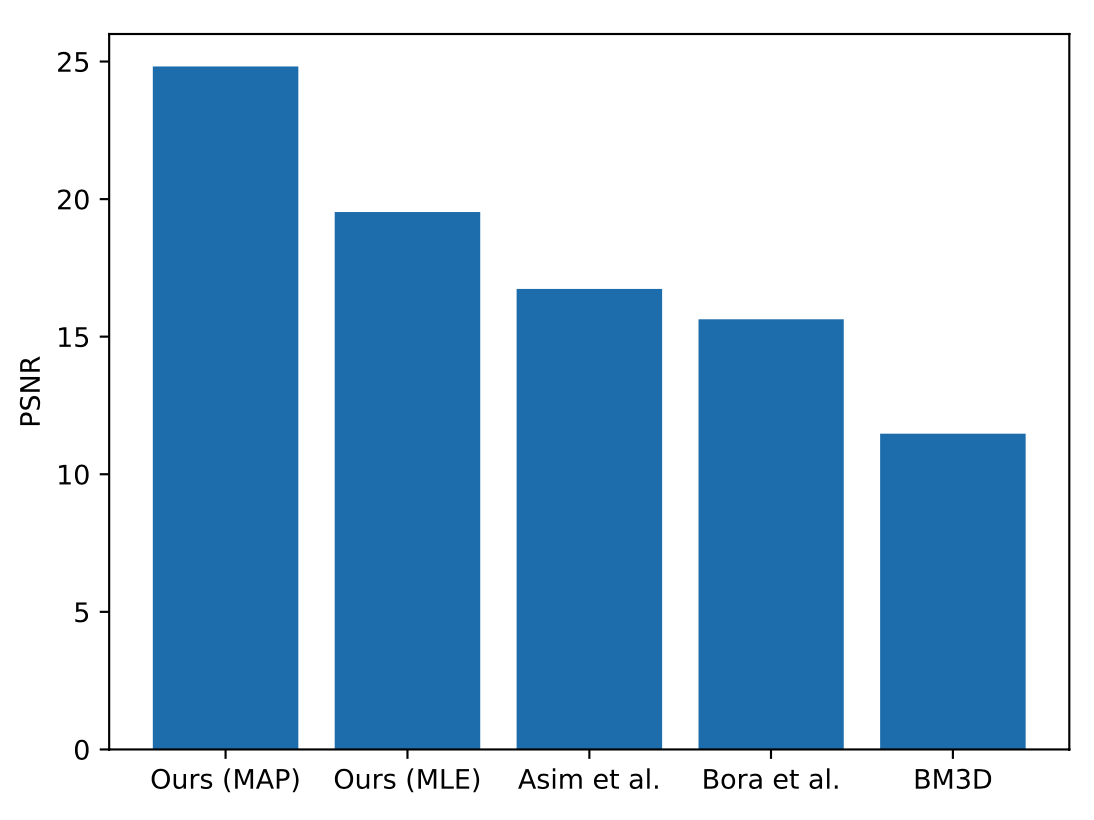}
\caption{Reconstruction PSNR on the MNIST denoising task.}
\label{fig:mnist_psnr}
\end{figure}

%% file: figures/fig_cs_noise_psnr_and_samples.tex
\begin{figure*}[!ht]
    \centering
    \begin{subfigure}[t]{0.45\linewidth}
        \centering
        \includegraphics[width=\textwidth]{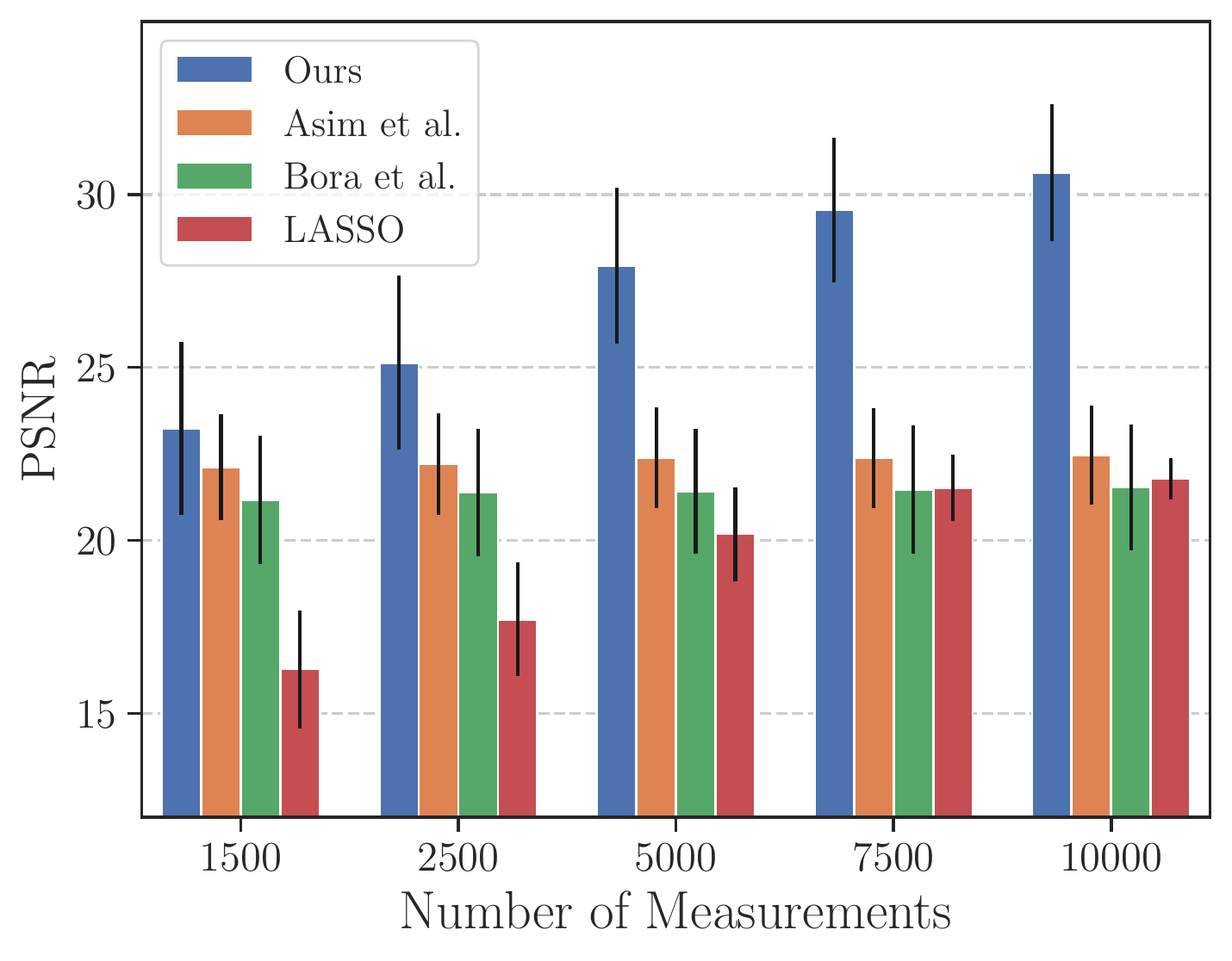}
        \vspace{-0.5em}
        \caption{
        PSNR at different measurement counts (best viewed in color). The approaches by \citet{PaulHandFlow} and \citet{bora2017compressed} show little improvements from having more measurements
        due to their inability to utilize the noise model.
        }
        \label{fig:cs_noise_psnr}
    \end{subfigure}
    \hfill
    \begin{subfigure}[t]{0.42\linewidth}
        \centering
        \includegraphics[width=\textwidth]{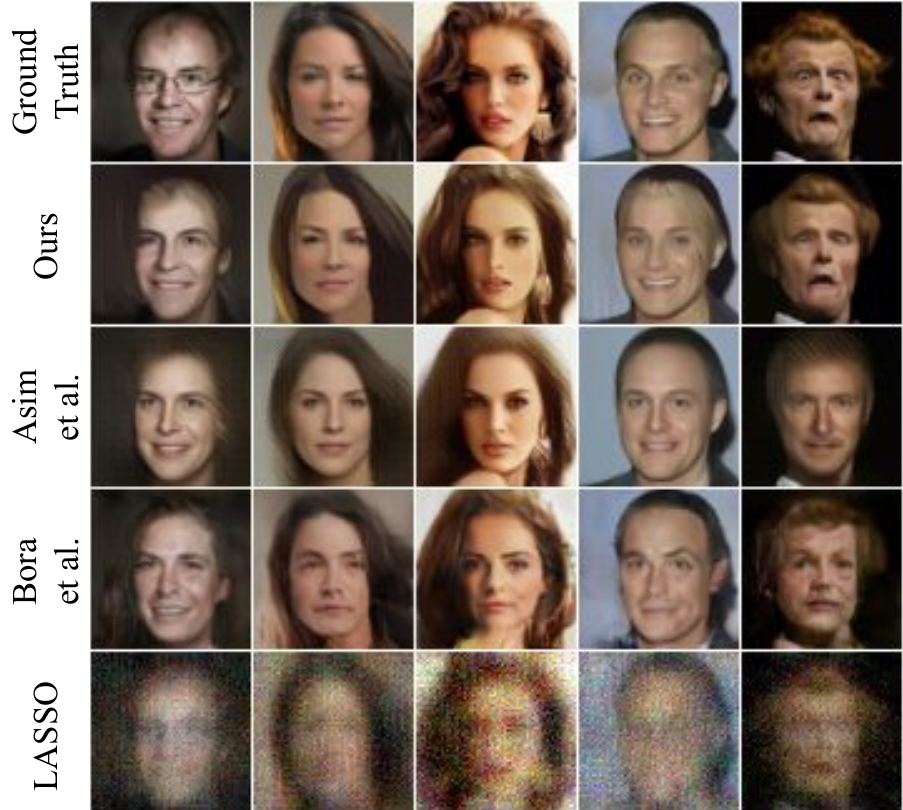}
        \vspace{-0.5em}
        \caption{
        Reconstructions at 2500 measurements.
        Notice that even though existing approaches produce reconstructions that resemble human faces,
        they do not match the ground truth as well as our method.
        }
        \label{fig:cs_noise_samples}
    \end{subfigure}
    \vspace{-0.5em}
    \caption{Experiment results for noisy compressed sensing on CelebA-HQ images.}
\end{figure*}

%% file: figures/fig_sinusoidal_samples_both.tex
\begin{figure*}[!ht]
\centering
\includegraphics[width=0.9\linewidth]{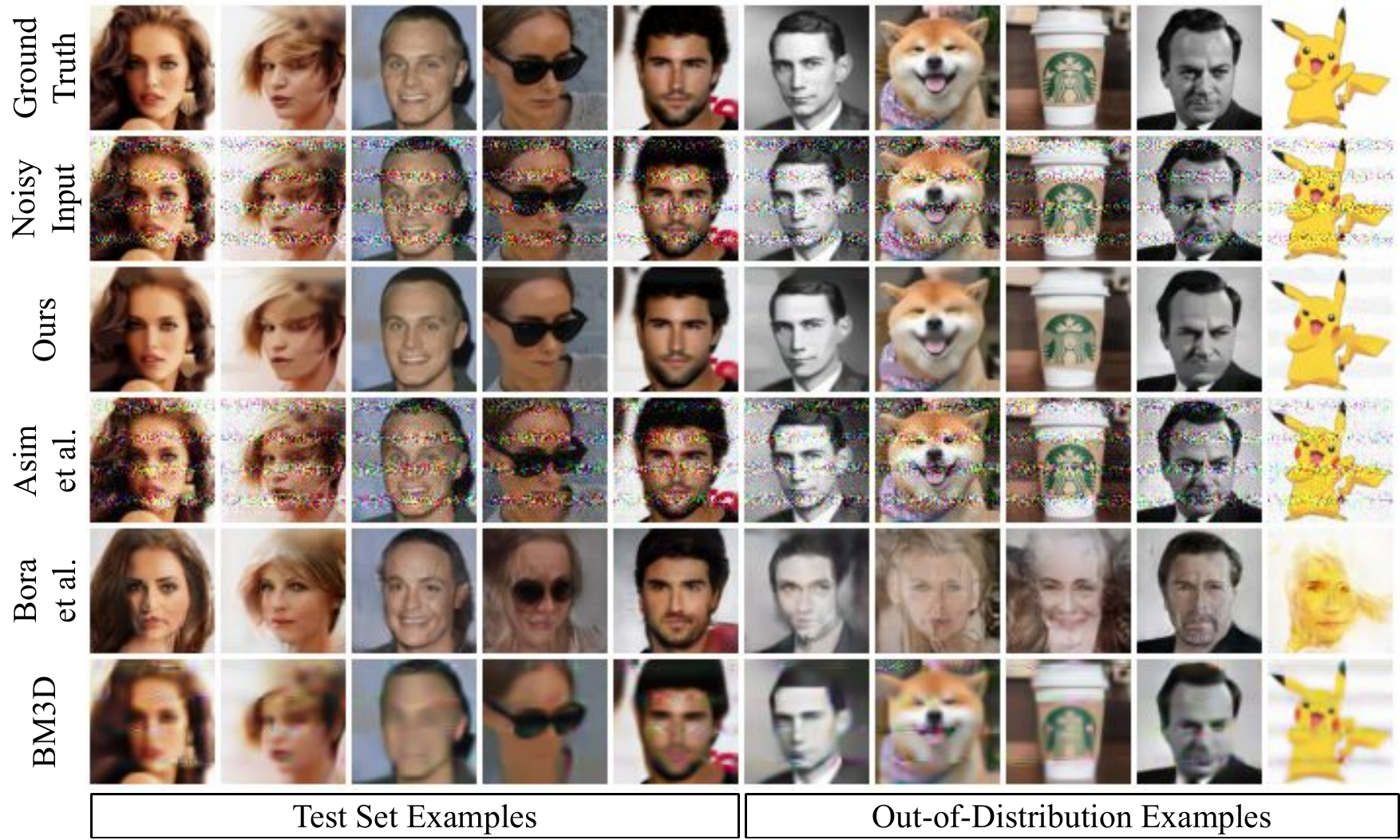}
\vspace{-0.8em}
\caption{Result of denoising \sinusoidal noise on CelebA-HQ faces and out-of-distribution images.}
\vspace{-0.5em}
\label{fig:sinusoidal_samples_both}
\end{figure*}

%% file: figures/fig_sinusoidal_and_hparam_sensitivity.tex
\begin{figure*}[!ht]
    \centering
    \begin{subfigure}[t]{0.48\textwidth}
        \centering
        \includegraphics[width=\textwidth]{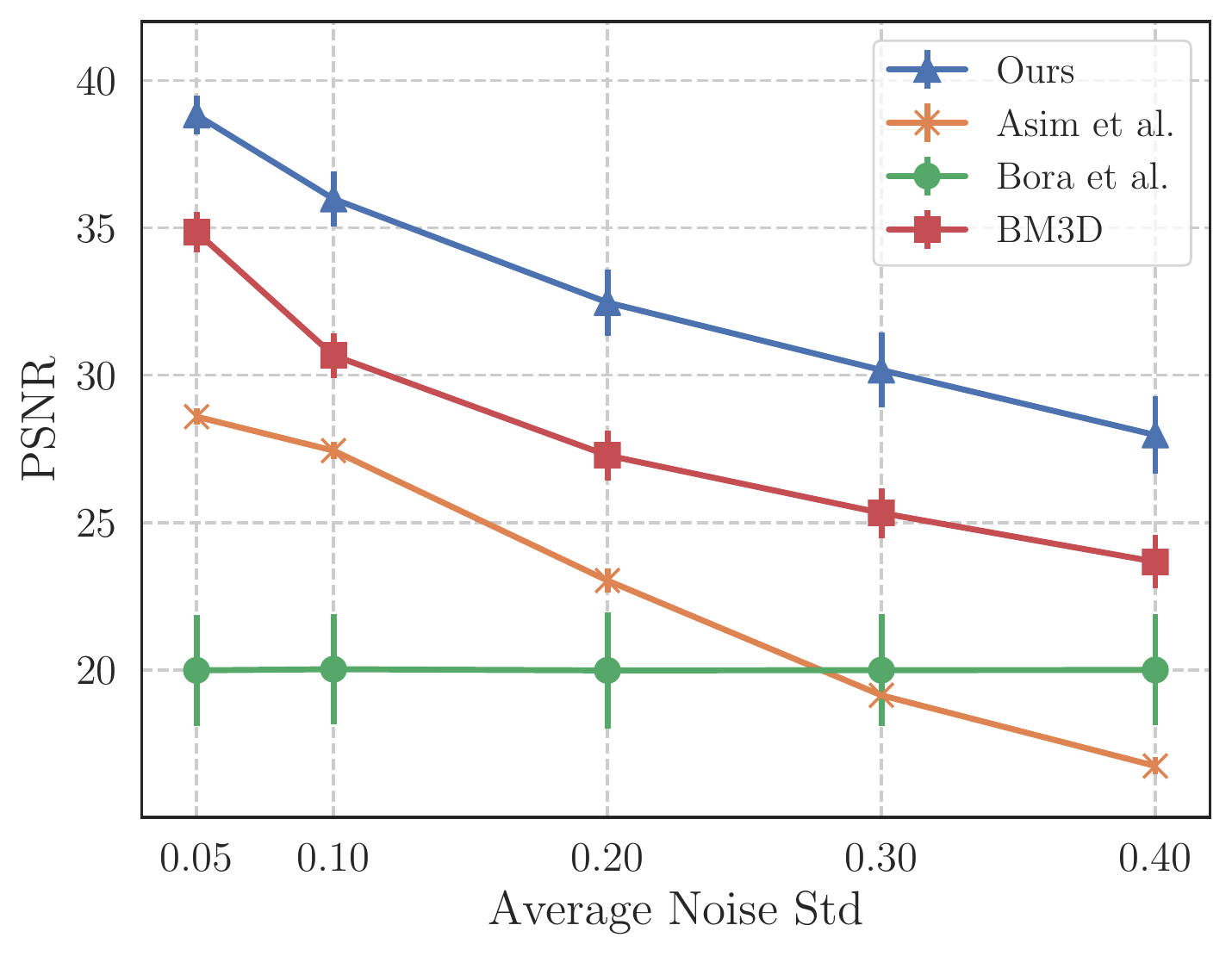}
        \caption{
        Result for \sinusoidal denoising on CelebA-HQ images at various noise rates. Our method achieves the same reconstruction performance even when the noise has up to $3\times$ higher average standard deviation
        compared to the best baseline method (BM3D).
        }
        \label{fig:sinusoidal_psnr}
    \end{subfigure}
    \hfill
        \begin{subfigure}[t]{0.48\textwidth}
        \centering
        \includegraphics[width=\textwidth]{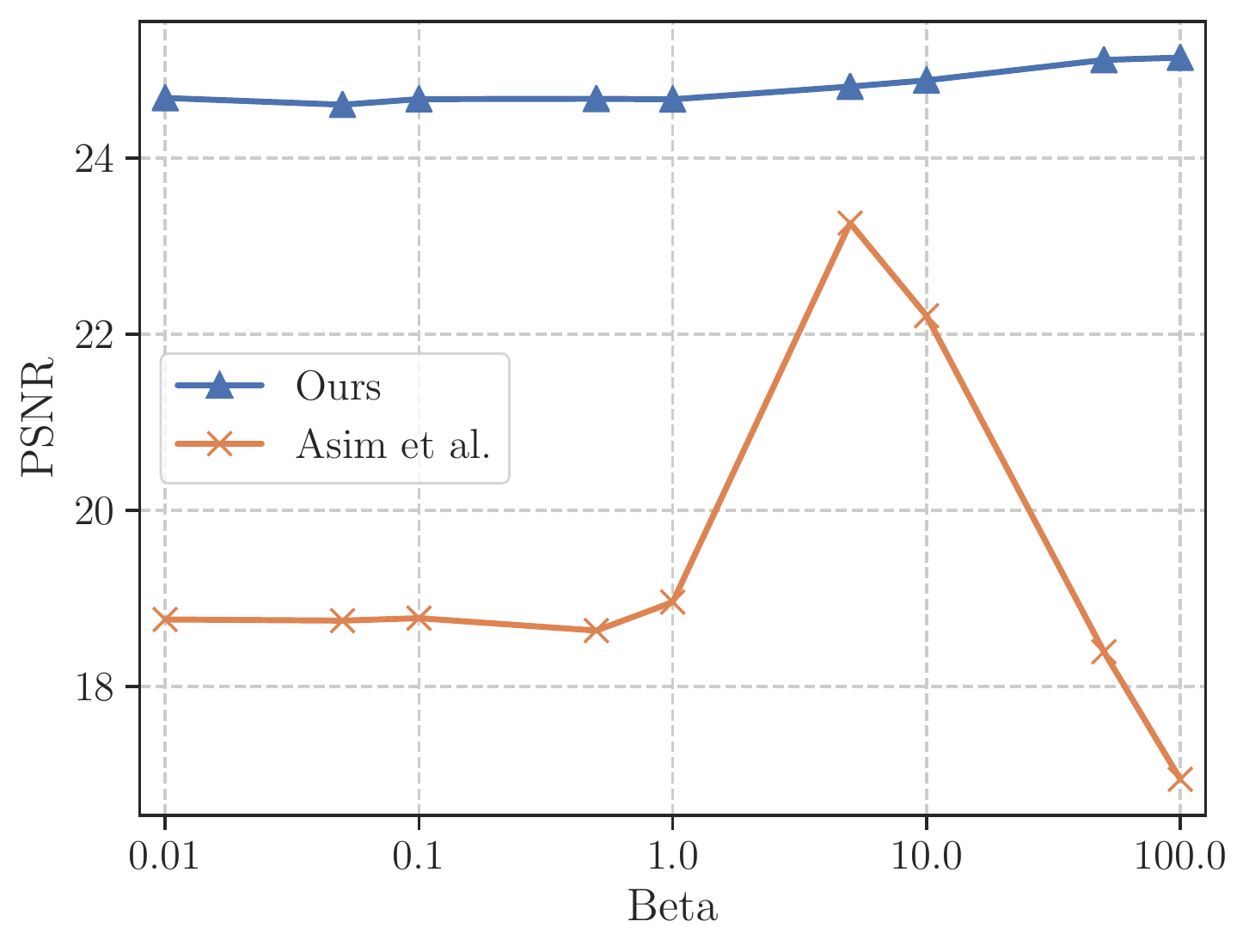}
        \caption{Compressed sensing performance of our method and \citep{PaulHandFlow} at different hyperparameter values.  For our method, we vary the smoothing parameter $\beta$ in $\LG$. For \citet{PaulHandFlow}, we vary the regularization coefficient $\gamma$ in $\Lhand$.}
        \label{fig:cs_noise_hparam_sensitivity}
    \end{subfigure}
    \caption{\sinusoidal denoising results (left) and hyperparameter sensitivity plot (right).}
    \vspace{-1em}
\end{figure*}

%% file: figures/fig_cs_1bit_psnr_and_samples.tex
\begin{figure*}[!ht]
    \centering
        \begin{subfigure}[t]{0.50\linewidth}
        \centering
        \includegraphics[width=\textwidth]{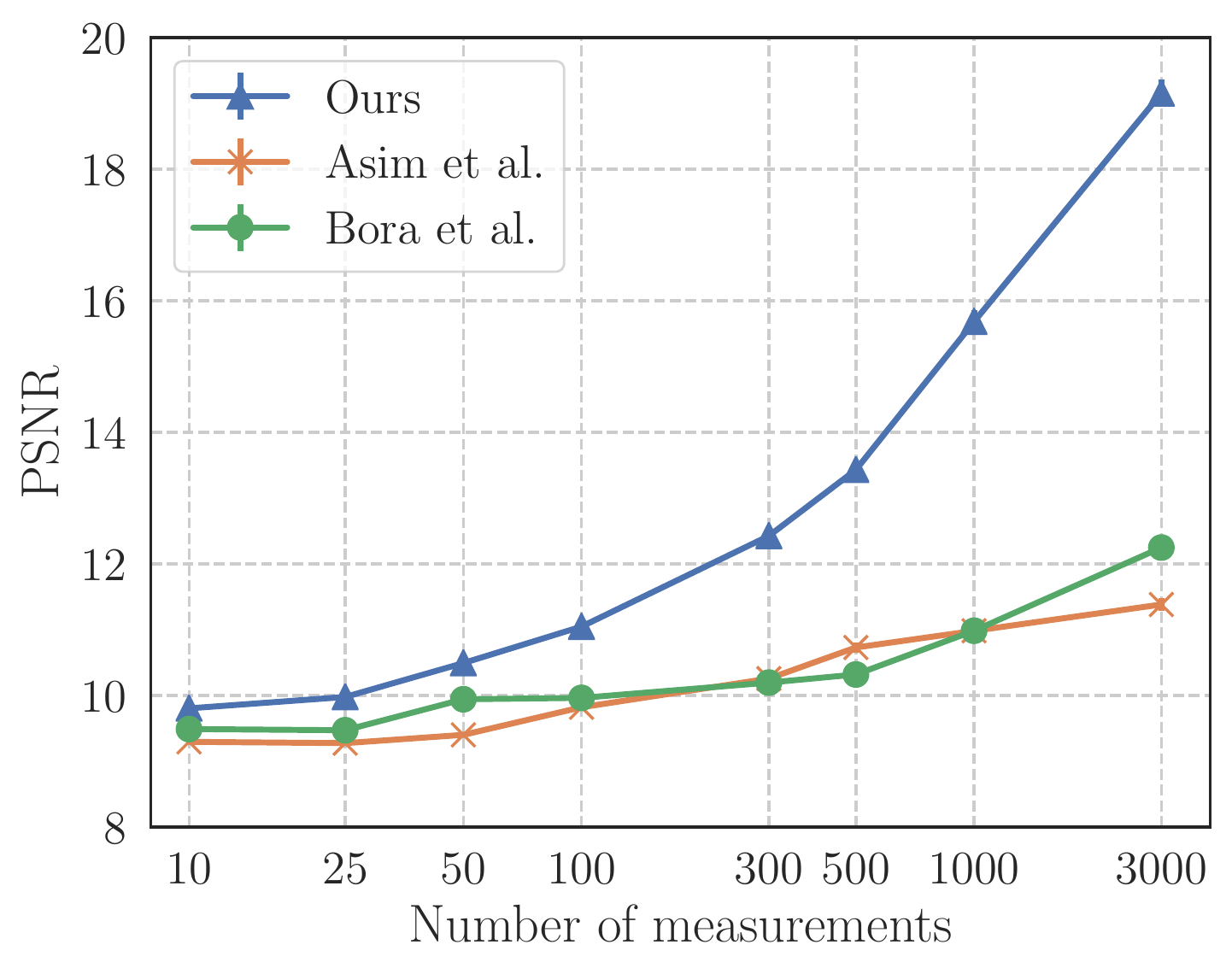}
        \caption{
        Result of 1-bit noisy compressed sensing at different measurement counts. Our method achieves the same reconstruction performance using up to $2\times$ fewer measurements compared to the best baseline method \citep{PaulHandFlow}.
        }
        \label{fig:cs_1bit_measurements}
    \end{subfigure}
    \hfill
    \begin{subfigure}[t]{0.44\textwidth}
        \centering
        \includegraphics[width=\columnwidth]{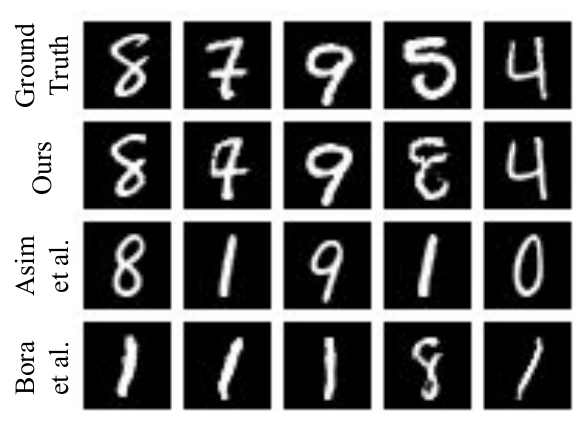}
        \caption{Reconstructions from noisy 1-bit compressed sensing with 3000 binary measurements.
        Notice that our method fails more gracefully compared to other methods, i.e. even when the reconstructions differ from the ground truth, substantial parts of the reconstructions are still correct.  On the other hand, other methods predict a completely different digit.
        }
        \label{fig:cs_1bit_samples}
    \end{subfigure}
    \caption{Results of 1-bit compressed sensing experiments.}
\end{figure*}

%% file: appendix.tex
\section{Omitted Proof}
\label{sec:proofs}

\subsection{Proof for Denoising}
\begin{proof}[Proof of \Cref{thm:denoising}]
We first show that gradient descent with sufficiently small learning rate will converge to $\bar{\bx}$, the locally-optimal solution of \Cref{eqn:denoising}. 
Recall the loss function $L(\bx) := q(\bx)+\frac{1}{2\sigma^2}\|\bx-\by\|^2$ (we subsume the scaling $\frac{1}{2}$ into $\frac{1}{\sigma^2}$ without loss of generality). Notice in the ball $B_{r}^d(\bx^*):=\set{\bx \in \R^d \,\vert\, \|\bx-\bx^*\|\leq r }$, $L$ is $\paren{\mu+\frac{1}{\sigma^2}}$ strongly-convex. We next show there is a stationary point $\bar{\bx} \in B_r^d(\bx^*)$ of $L(\bx)$.  
\begin{align*}
\nabla L(\bar{\bx})=0 & \Longrightarrow \nabla q(\bar{\bx})+\frac{1}{\sigma^2}(\bar{\bx}-\by)=0 \\
& \Longrightarrow \nabla q(\bar{\bx}) - \nabla q(\bx^*) = \frac{1}{\sigma^2}(\by - \bar{\bx}) \\
& \Longrightarrow \langle  \nabla q(\bar{\bx}) - \nabla q(\bx^*), \bar{\bx}-\bx^*\rangle \\
&= \frac{1}{\sigma^2}\langle \by - \bar{\bx}, \bar{\bx}-\bx^*\rangle
\end{align*}

From strong convexity of $q$,
$$\langle \nabla q(\bar{\bx}) - \nabla q(\bx^*), \bar{\bx}-\bx^*\rangle\geq \mu \|\bar{\bx}-\bx^* \|^2.$$
Thus,
\begin{align*}
\frac{1}{\sigma^2} &\langle \by - \bx^*,\bar{\bx}-\bx^*\rangle \\
&= \frac{1}{\sigma^2} \langle \paren{\by - \xbar}+\paren{\xbar-\bx^*},\bar{\bx}-\bx^*\rangle \\
&= \frac{1}{\sigma^2} \langle \by - \xbar,\bar{\bx}-\bx^*\rangle 
+ \frac{1}{\sigma^2} \langle \xbar-\bx^*,\bar{\bx}-\bx^*\rangle \\
&= \langle \nabla q(\bar{\bx}) - \nabla q(\bx^*), \bar{\bx}-\bx^*\rangle
+ \frac{1}{\sigma^2}\|\bar{\bx}-\bx^*\|^2 \\
&\geq \mu \|\bar{\bx}-\bx^* \|^2 + \frac{1}{\sigma^2}\|\bar{\bx}-\bx^*\|^2 \\
&= \paren{\mu+\frac{1}{\sigma^2}} \|\bar{\bx}-\bx^*\|^2
\end{align*}

Finally, by Cauchy-Schwartz inequality,
$$\langle \by - \bx^*,\bar{\bx}-\bx^*\rangle \leq \|\by - \bx^*\| \cdot \|\bar{\bx}-\bx^* \|.$$
So we get $\|\bar{\bx}-\bx^*\|\leq \frac{1}{1+\mu\sigma^2}\|\by-\bx^*\| \leq \|\bdelta\| \leq r$, in other words, $\bar{\bx}\in B_r^d(\bx^*)$.

Notice $L$ is $\paren{\mu+\frac{1}{\sigma^2}}$ strongly-convex in $B_r^d(\bx^*)$, which contains the stationary point $\bar{\bx}$. Therefore $\bar{\bx}$ is a local minimizer of $L(\bx)$. Also note that we implicitly require $q$ to be twice differentiable, meaning in a compact set $B_r^d(\bx^*)$ its smoothness is upper bounded by a constant $M$.  Thus gradient descent starting from $\by\in B_r^d(\bx^*)$ with learning rate smaller than $\frac{1}{M}$ will converge to $\bar{\bx}$ without leaving the (convex) set $B_r^d(\bx^*)$.  
\end{proof}

\section{Additional Experimental Results}
Here we include experimental results and details not included in the main text.
Across all the experiments, we individually tuned the hyperparameters for each method.

\subsection{Experimental Details}
\textbf{Dataset. }
For MNIST, we used the default split of 60,000 training images and 10,000 test images of \cite{lecun1998gradient}.
For CelebA-HQ, we used the split of 27,000 training images and 3,000 test images as provided by \cite{kingma2018glow}.

During evaluation, the following Python script was used to select 1000 MNIST images and 100 CelebA-HQ images from their respective test sets:
\begin{verbatim}
  np.random.seed(0)
  indices_mnist = np.random.choice(
      10000, 1000, False)
  np.random.seed(0)
  indices_celeba = np.random.choice(
      3000, 100, False)
\end{verbatim}
Note that CelebA-HQ images were further resized to $64\times64$ resolution.

\textbf{Noise Distributions. }
For the sinusoidal noise used in the experiments, the standard deviation of the $k$-th pixel/row is calculated as:
$$\sigma_k = 0.1 \cdot \paren{\exp\paren{\sin(2\pi\cdot\frac{k}{16})}-1} / (e-1),$$
clamped to be in range $[0.001, 1]$. For \Cref{fig:cs_1bit_scales}, we used vary the coefficient $0.1$ to values in $\set{0.05, 0.1, 0.2, 0.3, 0.4}$.

For the radial noise used in the additional experiment below, the standard deviation of each pixel with $\ell_2$ distance is $d$ from the center pixel (31, 31) is computed as:
$\sigma_k = 0.1 \cdot \exp(-0.005 \cdot d^2)$, clamped to be in range $[0.001, 1000]$.

\subsection{Additional Result: Removing \radial Noise}
Consider the measurement process $\xtil = \bx + \bdelta_{\text{radial}}$, where each pixel follows a Gaussian distribution, but with variance that decays exponentially in distance to the center point. For a pixel whose $\ell_2$ distance to the center pixel is $d$, the standard deviation is computed as $\sigma(d) = \exp\paren{-0.005\cdot d^2}$.
See \Cref{fig:radial_samples_both} and \Cref{fig:radial_psnr} for reconstructions as well as PSNR plot comparing the methods considered.

\input{figures/fig_radial_samples_both.tex}
\input{figures/fig_radial_psnr_and_cs_1bit_scales}

\subsection{Additional Result: 1-bit Compressed Sensing}
\Cref{fig:cs_1bit_scales} shows the performance of each method at different noise scales for a fixed number of measurements.  We observe that our method performs consistently better at all noise levels.

\section{Model Architecture and Hyperparameters}
\label{sec:models}
For the RealNVP models we trained, we used multiscale architecture as was done in \citep{dinh2016density}, with residual networks and regularized weight normalization on convolutional layers.  Following \citep{kingma2018glow}, we used 5-bit color depth for the CelebA-HQ model.  
Hyperparameters and samples from the models can be found in \Cref{table:realnvp_hparams} and \Cref{fig:flow_images}.
\input{figures/realnvp_hparams.tex}
\input{figures/realnvp_samples.tex}
\input{figures/ood_images.tex}

\section{Experiment Hyperparameters}
Here we list the hyperparameters used for each experiment.  We used the Adam optimizer \citep{kingma2014adam} for all appropriate methods below.

\textbf{Denoising MNIST Digits.}
\begin{itemize}[topsep=0pt,itemsep=2pt,partopsep=0pt,parsep=0pt,leftmargin=5mm]
  \item Learning rate: $0.02$
  \item Optimization steps for Ours (MAP) and \citep{PaulHandFlow}: $400$
  \item Optimization steps for Ours (MLE) and \citep{bora2017compressed}: $1000$
  \item Smoothing parameter for Ours (MAP \& MLE): $\beta = 1.0$
  \item Regularization for \citep{PaulHandFlow}: $\gamma = 0.0$
  \item Regularization for \citep{bora2017compressed}: $\lambda = 0.01$
\end{itemize}

\textbf{Noisy Compressed Sensing. }
\begin{itemize}[topsep=1pt,itemsep=2pt,partopsep=0pt,parsep=0pt,leftmargin=5mm]
  \item Learning rate: $0.02$
  \item Optimization steps for Ours (MAP) and \citep{PaulHandFlow}: $300$
  \item Optimization steps for \citep{bora2017compressed}: $1000$
  \item Smoothing parameter for Ours (MAP): $\beta = 100$
  \item Regularization for \citep{PaulHandFlow}: $\gamma = 10$
  \item Regularization for \citep{bora2017compressed}: $\lambda = 0.001$
  \item Regularization for LASSO: $\lambda = 0.01$
\end{itemize}

\textbf{Denoising Sinusoidal Noise. }
\begin{itemize}[topsep=1pt,itemsep=2pt,partopsep=0pt,parsep=0pt,leftmargin=5mm]
  \item Learning rate: $0.02$
  \item Optimization steps for Ours (MAP) and \citep{PaulHandFlow}: $150$
  \item Optimization steps for \citep{bora2017compressed}: $1000$
  \item Smoothing parameter for Ours (MAP): $\beta = 0.5$
  \item Regularization for \citep{PaulHandFlow}: $\gamma = 2.0$
  \item Regularization for \citep{bora2017compressed}: $\lambda = 0.01$
\end{itemize}

\textbf{Noisy 1-bit Compressed Sensing. }
\begin{itemize}[topsep=1pt,itemsep=2pt,partopsep=0pt,parsep=0pt,leftmargin=5mm]
  \item Learning rate: $0.02$
  \item Optimization steps for Ours (MAP) and \citep{PaulHandFlow}: $200$
  \item Optimization steps for \citep{bora2017compressed}: $1000$
  \item Smoothing parameter for Ours (MAP): $\beta = 1.0$
  \item Regularization for \citep{PaulHandFlow}: $\gamma = 1.0$
  \item Regularization for \citep{bora2017compressed}: $\lambda = 0.01$
\end{itemize}

%% file: figures/fig_radial_samples_both.tex
\begin{figure*}[!ht]
\centering
\includegraphics[width=0.9\linewidth]{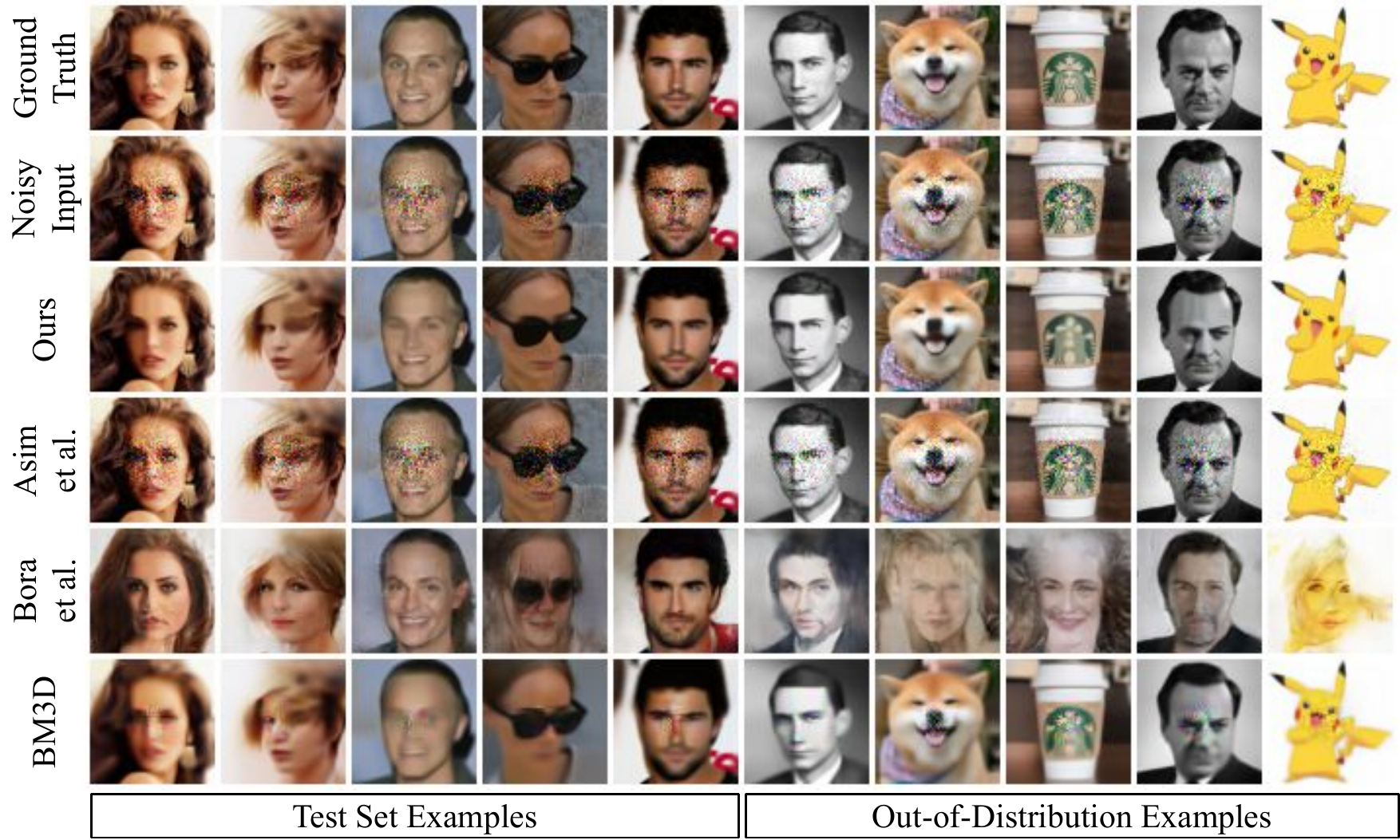}
\caption{Result of denoising \radial noise on CelebA-HQ faces and out-of-distribution images.}
\label{fig:radial_samples_both}
\end{figure*}

%% file: figures/fig_radial_psnr_and_cs_1bit_scales.tex
\begin{figure*}[!ht]
    \centering
    \begin{subfigure}[t]{0.48\textwidth}
        \centering
        \includegraphics[width=\textwidth]{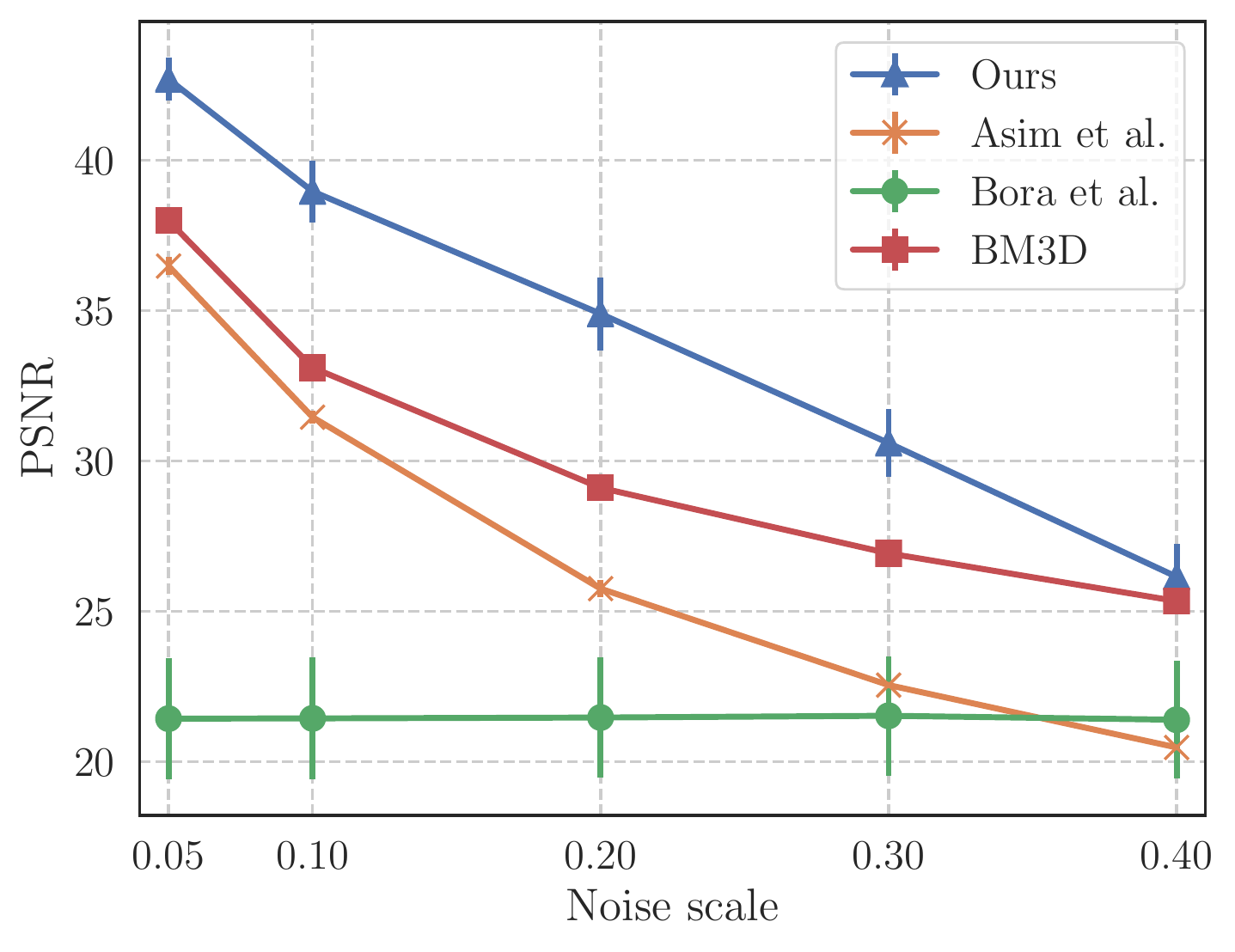}
        \caption{Result on denoising \radial noise at varying noise rates. Our method achieves the same reconstruction performance even when the noise has approximately $1.5\times$ higher noise scale compared to the best baseline method which is BM3D for this setting.
        }
        \label{fig:radial_psnr}
    \end{subfigure}
    \hfill
        \begin{subfigure}[t]{0.48\textwidth}
        \centering
        \includegraphics[width=\textwidth]{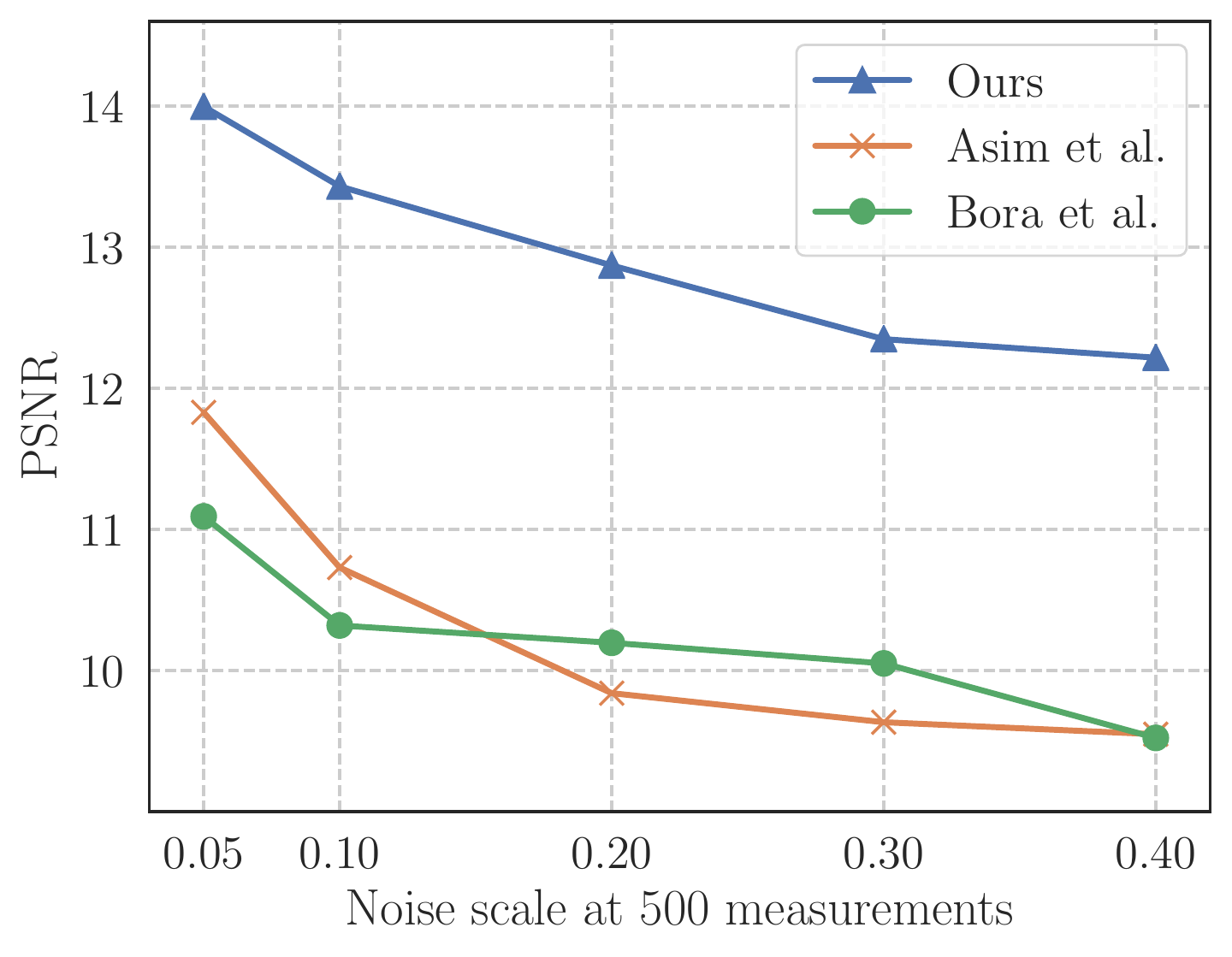}
        \caption{Result of 1-bit compressed sensing at different noise scale. Our method obtains the best reconstructions, achieving similar PSNR as \citep{PaulHandFlow} when the noise scale is $8 \times$ higher.}
        \label{fig:cs_1bit_scales}
    \end{subfigure}
    \caption{\radial denoising results (left) and 1-bit compressed sensing results at different noise levels (right).}
\end{figure*}

%% file: figures/realnvp_hparams.tex
\begin{table}[!ht]
\centering
\begin{tabular}{l|c|c}
\textbf{Hyperparameter} & \textbf{CelebA-HQ} & \textbf{MNIST} \\ \hline
Learning rate           & $5\mathrm{e}{-4}$ & $1\mathrm{e}{-3}$ \\
Batch size              & 16                        & 128 \\
Image size              & $64 \times 64 \times 3$   & $28 \times 28 \times 1$ \\
Pixel depth             & 5 bits                    & 8 bits \\
Number of epochs        & 300                       & 200 \\
Number of scales        & 6                         & 3 \\
Residual blocks per scale & 10                        & 6 \\
Learning rate halved every & 60 epochs              & 40 epochs  \\
Max gradient norm       & 500                       & 100 \\
Weightnorm regularization   & $1\mathrm{e}{-5}$         & $5\mathrm{e}{-5}$ \\
\end{tabular}
\vspace{1em}
\caption{Hyperparameters used for RealNVP models.} 
\label{table:realnvp_hparams}
\end{table}

%% file: figures/realnvp_samples.tex
\begin{figure}[!ht]
\centering
\includegraphics[width=0.9\linewidth]{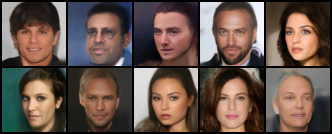}
\vspace{1em}
\includegraphics[width=0.9\linewidth]{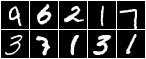}
\caption{Samples from the RealNVP models used in our experiments.}
\label{fig:flow_images}
\end{figure}

%% file: figures/ood_images.tex
\begin{figure}[!ht]
\centering
\includegraphics[width=\linewidth]{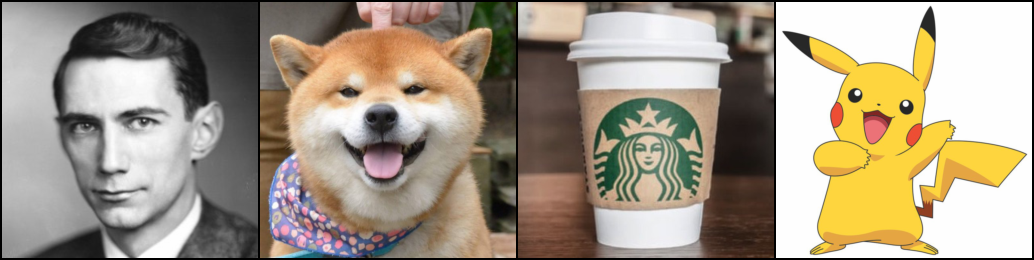}
\caption{Out-of-distribution images used in our experiments. We included different types of out-of-distribution instances including grayscale images and cartoons with flat image areas. }
\label{fig:ood_images}
\end{figure}